\documentclass{article}

\usepackage{arxiv}

\usepackage[utf8]{inputenc} 
\usepackage[T1]{fontenc}    
\usepackage{hyperref}       
\usepackage{url}            
\usepackage{booktabs}       
\usepackage{amsfonts}       
\usepackage{nicefrac}       
\usepackage{microtype}      
\usepackage{lipsum}
\usepackage{graphicx}
\usepackage[figuresright]{rotating}
\usepackage[section]{placeins}
\usepackage{amsmath}
\usepackage{float}
\graphicspath{ {./images/} }
\usepackage{epstopdf}
\title{Multi-Viewpoint and Multi-Evaluation with Felicitous Inductive Bias Boost Machine Abstract Reasoning Ability}

\author{
Qinglai Wei\\
State Key Laboratory for Management and Control of Complex Systems, \\Institute of Automation, Chinese Academy of Sciences\\
School of Artificial Intelligence, University of Chinese Academy of Sciences\\
Beijing, China \\
\texttt{qinglai.wei@ia.ac.cn}\\
\And
Diancheng Chen\\
State Key Laboratory for Management and Control of Complex Systems, \\Institute of Automation, Chinese Academy of Sciences\\
School of Artificial Intelligence, University of Chinese Academy of Sciences\\
Beijing, China \\
\texttt{chendiancheng2020@ia.ac.cn}\\
\And
Beiming Yuan\\
School of Artificial Intelligence, University of Chinese Academy of Sciences\\
Beijing, China\\
\texttt{yuanbeiming20@mails.ucas.ac.cn}\\
}

\begin{document}
\maketitle
\footnote{All authors contributed equally to this work.}
\footnote{Corresponding Author: Diancheng Chen (chendiancheng2020@ia.ac.cn)}
\begin{abstract}
Great endeavors have been made to study AI's ability in abstract reasoning, along with which different versions of RAVEN's progressive matrices (RPM) are proposed as benchmarks. Previous works give inkling that without sophisticated design or extra meta-data containing semantic information, neural networks may still be indecisive in making decisions regarding the RPM problems, after relentless training. Evidenced by thorough experiments, we show that, neural networks embodied with felicitous inductive bias, intentionally design or serendipitously match, can solve the RPM problems efficiently, without the augment of any extra meta-data. Our work also reveals that multi-viewpoint with multi-evaluation is a key learning strategy for successful reasoning. Nevertheless, we also point out the unique role of meta-data by showing that a pre-training model supervised by the meta-data leads to a RPM solver with better performance. Source code can be found in https://github.com/QinglaiWeiCASIA/RavenSolver.
\end{abstract}


\section{Introduction}\label{introduction}

From expert system with elaborately designed rules to the renaissance of neural network, AI practitioners never cease to work on machine intelligence to make it a counterpart of human intelligence. The tremendous success of machine learning in areas like visual perception \cite{ImageNet,AlexNet,ResNet}, natural language processing \cite{Transformer, Bert, GPT-3}, or generative models \cite{GAN,VAE,DiffusionModel} , intrigues researchers to study the reasoning ability of AI. Representative works cover, but not limit to, visual question answering \cite{VQA,CLEVERdataset}, flexible application of language models \cite{Math1,Math2,CodeX}, and abstract reasoning problems \cite{Bongard1,Bongard2}. Here we consider the RPM problem, originally develops for the purpose of IQ test \cite{RPM}, and recently serves as a benchmark for the evaluation of AI's abstract reasoning ability.

\begin{figure}[h]%
\centering
\includegraphics[width=0.6\textwidth]{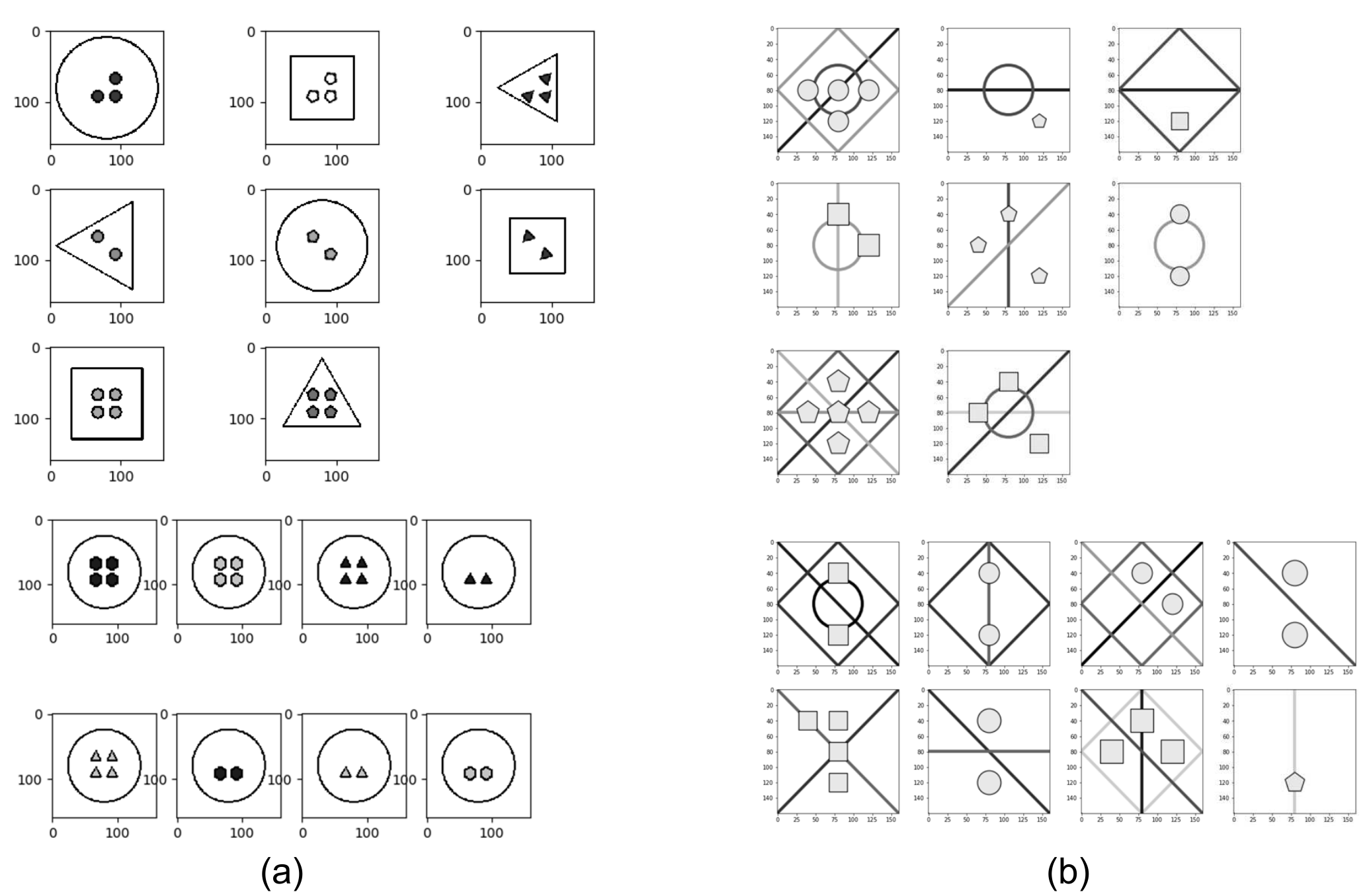}
\caption{Demonstrations of RPM problems. These two RPM questions are snapshots from I-RAVEN and PGM dataset, respectively.}
\label{RPM_Demo}
\end{figure}

Fig. \ref{RPM_Demo} shows two RPM problems. Without loss of generality, RPM problems are formalized within three steps. First, sample rules which determine the changing patterns of visual attributes, from a predefined rule set. Common rules include, but not limited to, arithmetic operation, set operation, and logic operation. Second, given the sampled rules, design proper values for all the visual attributes. Common visual attributes are type, size, and color, etc. Some visual attributes may play the role of distracter, with their values change randomly. Finally, render images basing on all the visual attribute values. Instantiated RPM problem is composed of a context and an answer pool: the context is a 3 $\times$ 3 image matrix, with image in the lower right corner missing. While the answer pool contains 8 images for selection, and the test-takers are expected to select one most fitted image from the answer pool to complete the matrix, so as to make it compatible with the internal rules.

To achieve satisfying reasoning accuracy in RPM problems, it is expected that models should be able to extract visual attributes relevant to the downstream tasks, in the meantime infer about the underlying rules. That is, traditional perception neural networks consisting of perception modules only is incompetent to solve the RPM problems \cite{RAVENdataset, PGMdataset}.

In this work, we solve the RPM problems in an end-to-end manner. Several key points to follow when developing the black-box RPM solver: distinct modularization to imitate the complete perception and reasoning processes, encapsulation of two potential RPM characteristics, namely permutation-invariance and transpose-invariance, into the inductive bias design, and the implementation of multi-viewpoint and multi-evaluation strategy. To be specific, distinct modularization requires both the cooperation and a clear boundary between the feature extraction module and the reasoning module. It is expected that each module attends to its own duty properly, otherwise adding a new module is nothing but merely extending the depth of a neural network. This issue is addressed by injecting available inductive bias to the reasoning module to make it aware of the permutation-invariance and transpose-invariance characteristics of the RPM problems. On the other hand, various visual attributes and rules are involved in the RPM problems, resulting in abundant attribute-rule combinations. In light of this, we equip the feature extraction module with multi-viewpoint strategy and the reasoning module with multi-evaluation strategy, which endows with the ability of attending to the RPM problems in different perspectives to the model. Aforementioned details will suffice to build a RPM solver with very high reasoning accuracy. Nevertheless, we train a auxiliary model to predict the natural languages describing the rules for the RPM problems. Adopting this auxiliary model as a pre-training model, we manage to train a RPM solver with higher reasoning accuracy in a very fast manner.

The results of our work are promising and intriguing in several ways. First, it shows that, models with multi-viewpoint and multi-evaluation strategies, either based on convolutional neural network (CNN) or vision transformer (ViT \cite{ViT}), produce competitive reasoning accuracies, without the aid of any meta-data. Second, it is shown experimentally that rules captured by the neural network are different from the predefined rules. Third, we find out that model predicting the rules of the RPM problem can serve well as a pre-training model for the RPM solver, which bring forth higher reasoning accuracy and faster training speed.

\section{Related Work}\label{relatedwork}

\subsection{RPM Dataset}

We study RAVEN \cite{RAVENdataset}, I-RAVEN \cite{I-RAVEN}, and PGM \cite{PGMdataset} datasets in this work. All these datasets follow the general construction guideline described before, but they differ in subtle ways.

RAVEN consists of 7 distinct configurations with different difficulty levels. The easiest configuration is `Center', where each panel of the problem matrix only has one entity, while harder configurations such as `3 $\times$ 3 Grid' has at most nine entities in each panel. Test-takers are required to observe the changing patterns row-wise, extract visual attributes, summarize rules controlling the row-wise changes of visual attributes, then make choice to complete the problem matrix. The most difficult configuration, `O-IG', as shown in Fig. \ref{RPM_Demo}(a), requires test-takers to divide entities in each panel into two groups, each of which follows one set of rules, then perform reasoning respectively. Some literatures show that the answer generation process of RAVEN encourages neural network solvers to find shortcut solutions instead of discovering rules \cite{I-RAVEN, MRNet}, and datasets like I-RAVEN and RAVEN-Fair with refining answer generation strategies are proposed to address this issue \cite{I-RAVEN,MRNet}.

Fig. \ref{RPM_Demo}(b) shows an example of the PGM dataset, where each panel of the PGM problem matrix may have entities in the foreground and lines in the background. Test-takers need to observe the changes of visual attributes row-wise and column-wise simultaneously, summarize the potential rules in the foreground and background respectively, and then complete the reasoning task accordingly.

Statistically speaking, in average, RAVEN and I-RAVEN possesses more rules than PGM per question (6.29 vs. 1.37 \cite{RAVENdataset}). RAVEN and I-RAVEN has two fixed visual attributes as distractors, while PGM is way more flexible in that any visual attribute can be a distractor. Rules of RAVEN and I-RAVEN are encoded row-wise, while one must check row-wise and column-wise information simultaneously for summarizing rules in PGM.

\subsection{RPM solvers}

literatures of RPM solver expand rapidly in recent years. Here we roughly divide them into two categories. The first one is end-to-end black-box solvers, accounting for the majority of previous works. The second one leverages symbolic AI in order to obtain results beyond reasoning accuracies, such as interpretability.

The end-to-end black-box models focus on improving the reasoning accuracy on RPM problems. Early works show that prevalent visual models fail to solve RPM problems, and adding extra labels containing information of structure or rule improve the results to some extent \cite{RAVENdataset,PGMdataset}. In LEN\cite{LEN}, researchers argue that the main challenge in solving RPM problems is the elimination of distracting information. CoPINet \cite{CoPINet} and DCNet \cite{DCNet} are proposed to leverage contrastive learning in reasoning. MRNet \cite{MRNet} shows that retrieving features from different CNN blocks which connect serially helps the model to capture multiple visual attributes simultaneously, it is also the first work to report that extra meta-data jeopardizes network performance. In SCL \cite{SCL}, tensor scattering is performed to make each scattered part attend to specific visual attributes or rules. SAVIR-T \cite{SAVIR-T} extracts intra-image information and inter-image relations so as to facilitate reasoning ability.

Symbolic AI powered methods bring forth higher reasoning accuracies and stronger model interpretability. In PrAE \cite{PrAE}, a neural symbolic system performs probabilistic abduction and execution to generate an answer image. ALANS \cite{ALANS} manages to get rid of prior knowledge required in PrAE and outperforms monolithic end-to-end model in terms of generalization ability. NVSA \cite{NVSA} uses holographic vectorized representations and ground-truth attribute values to build a neural-symbolic model.

In one hand, our methods absorb successful experiences of previous models. Specifically, we fully utilize the inductive bias of the RPM problem like MRNet and SAVIR-T do, and adopt the encoder architecture of MRNet in one of our models. On the other hand, the active expressiveness of inductive bias in our models, and the unique multi-viewpoint and multi-evaluation strategies, make our models stand out from previous models, in terms of the reasoning accuracy.

\subsection{CLIP}

CLIP is a multi-modal pre-training neural network, which jointly trains an image encoder and a natural language encoder. By maximizing the similarity between the visual representation and natural language embedding of the positive sample pairs and minimizing the aforementioned similarity in the negative sample pairs, CLIP learns visual representations of high quality, which enables zero-shot transfer to downstream tasks \cite{Clip}.

In our study, we show that our model produces unaligned rule representations for RPM problem matrices with the same rule. To guide the behaviour of our model, we train a CLIP model with a specific mask scheme to align the rule representation of each RPM problem matrix with the embedding of natural language describing the corresponding rule, then regard the visual end of the trained CLIP as a pre-trained perception module for our model. As a result, we obtain a new model with remarkably high reasoning accuracies and fast convergence speed, compared with our original model without pre-training.

\section{Method}

Here we give the definition of the RPM problem: $\left\{ {X_{_{p}}^i} \right\}_{i = 1}^8$ denotes the ordered images in each $3 \times 3$ problem matrix, with the image in the lower right corner missing. $\left\{ {X_{_{ac}}^i} \right\}_{i = 1}^8$ denotes the unordered answer candidates. Test-taker is expected to select one image from the answer candidates to complete the problem matrix.

We first introduce our RPM solvers in two forms, namely RS-CNN and RS-TRAN, which are composed of convolutional neural networks and transformer blocks respectively. We show that RS-CNN can perform accurate reasoning in RAVEN and I-RAVEN datasets, with proper inductive bias design, while the inductive bias of RS-TRAN naturally lends itself to all the RPM problems without extra design, and that multi-viewpoint with multi-evaluation mechanism is able to improve the reasoning ability of RS-TRAN remarkably. Then we discuss the potential problems of the original meta-data, and introduce RS-TRAN-CLIP, which is a masked CLIP-based pre-training model for RS-TRAN.

\subsection{RS-CNN}

\begin{figure}[h]%
\centering
\includegraphics[width=0.8\textwidth]{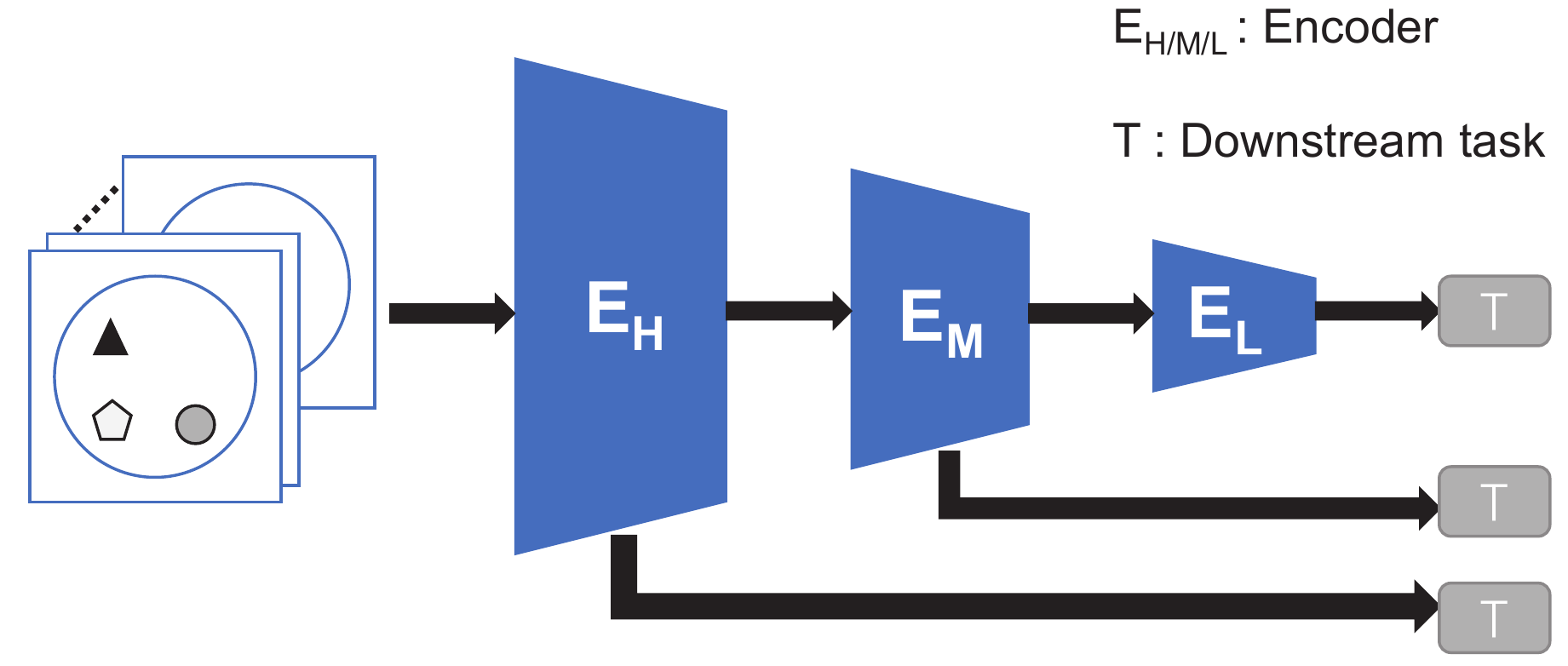}
\caption{Simple illustration of multi-scale encoder developed in MRNet. $E_H$, $E_M$, and $E_L$, serially connected, are residual convolutional blocks with decreasing kernel size. Not only the information processed by a former block will flow into the successor block, but also the output of each block will serve as representation for the downstream tasks individually.}
\label{CNN_Perception}
\end{figure}

RS-CNN consists of a perception module and a reasoning module. The perception module is expected to capture various visual attributes simultaneously. We follow the architecture of multi-scale encoder of MRNet\cite{MRNet}, with different convolutional blocks attending to different visual attributes, as shown in Fig. \ref{CNN_Perception}. For images in a problem matrix $\left\{ {X_{_{p}}^i} \right\}_{i = 1}^8$ and the corresponding answer candidate $\left\{ {X_{_{ac}}^i} \right\}_{i = 1}^8$, the perception module of RS-CNN produces representation triplets $\{ e_{p,h}^i,e_{p,m}^i,e_{p,l}^i\}_{i = 1}^8 $ and $\{ e_{ac,h}^i,e_{ac,m}^i,e_{ac,l}^i\}_{i = 1}^8 $, where $h, m, l$ refers to the convolutional blocks $E_H, E_M, E_L$ in Fig. \ref{CNN_Perception}, respectively.

\begin{figure}[h]%
\centering
\includegraphics[width=0.9\textwidth]{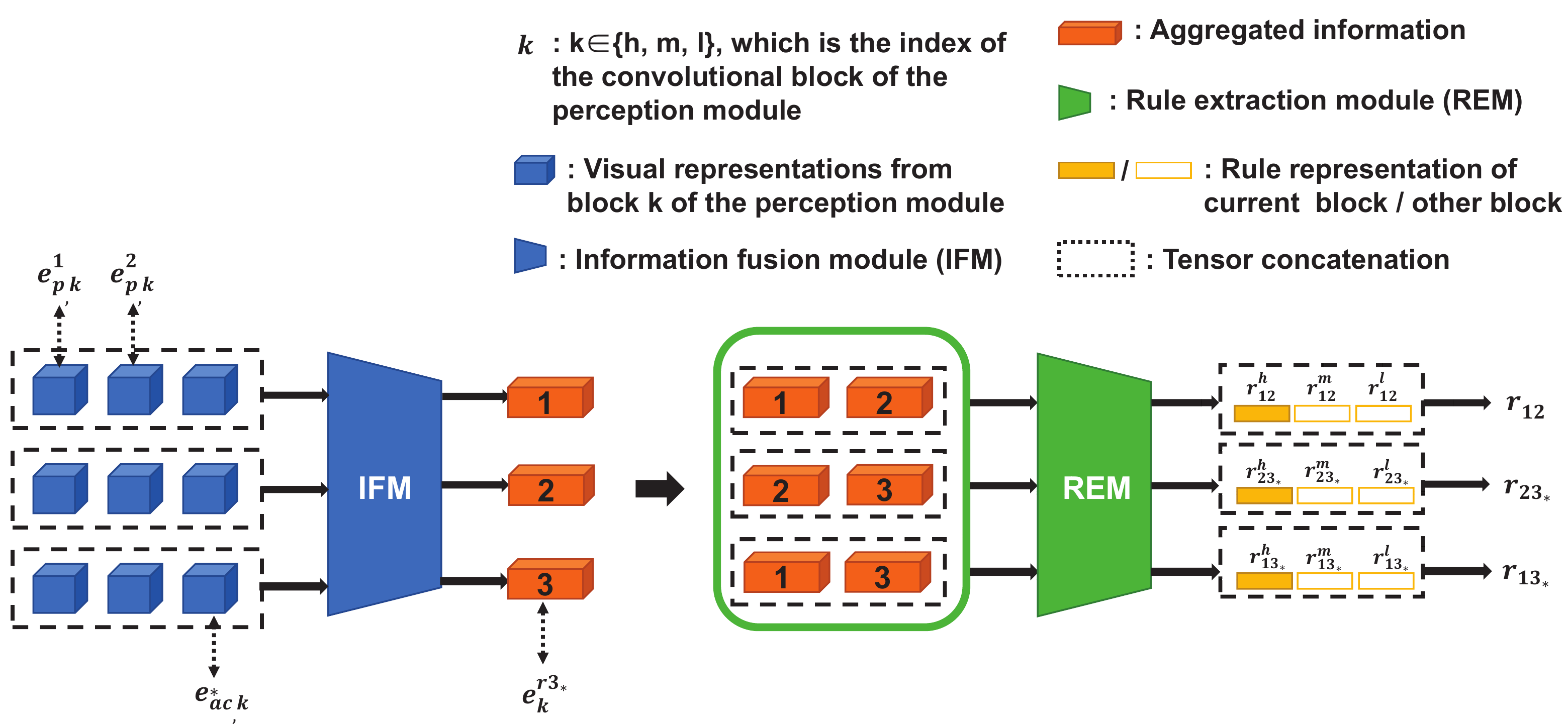}
\caption{Simple illustration of the reasoning module of RS-CNN. The reasoning module takes the features produced by each block of the perception module individually, and generates aggregated row information of the problem matrix by IFM, then summarizes potential rules for each pair of aggregated row information by REM, and finally concatenates the summarized rules of each block, in the order of $E_H$, $E_M$, and $E_L$.}
\label{CNN_Reasoning}
\end{figure}

Fig. \ref{CNN_Reasoning} depicts the reasoning module, which constitutes the above-mentioned downstream task $T$. The reasoning module takes outputs from each convolutional block of the perception module ($E_H$, $E_M$, and $E_L$) as inputs. Take $E_H$ as example, the representations of images in each problem matrix are concatenated row-wise and fed into an information fusion module (IFM) formed by by convolutional blocks to obtain the row-wise aggregated representations:

\begin{align}
row 1 :IFM[Concatenate\{ e_{p,h}^1,e_{p,h}^2,e_{p,h}^3\} ] = e_h^{r1}\\
row 2 :IFM[Concatenate\{ e_{p,h}^4,e_{p,h}^5,e_{p,h}^6\} ] = e_h^{r2}\\
row 3 :IFM[Concatenate\{ e_{p,h}^7,e_{p,h}^8,e_{ac,h}^*\} ] = e_h^{r3_*},
\end{align}
where $e_{ac,h}^* \in \{ e_{ac,h}^i\} _{i = 1}^8$.

Note that rules can only be determined after observing at least two rows of the problem matrix. Take the aggregated representations of $E_H$ as example, every two rows of aggregated information are paired up and fed into a rule extraction module (REM) formed by bottleneck residual convolutional blocks to obtain the rule representation, in the perspective of $E_H$:

\begin{align}
REM(e_h^{r1},e_h^{r2}) = r_{12}^h\\
REM(e_h^{r2},e_h^{r3_*}) = r_{23_*}^h\\
REM(e_h^{r1},e_h^{r3_*}) = r_{13_*}^h,
\end{align}

Finally, we concatenate the rules derived by each block of the perception module ($E_H, E_M, E_L$) to be the final rule representation:

\begin{align}
Concatenate(r_{12}^h,r_{12}^m,r_{12}^l) = {r_{12}}\\
Concatenate(r_{23_*}^h,r_{23_*}^m,r_{23_*}^l) = {r_{23_*}}\\
Concatenate(r_{13_*}^h,r_{13_*}^m,r_{13_*}^l) = {r_{13_*}},
\end{align}

The network architecture introduced so far covers all the basic elements for constructing RS-CNN. Apart from the perception module, the aggregated information and observing two rows simultaneously for rule extraction resembles triplets in the relation module of MRNet and the shared rule extraction mechanism in SAVIR-T, and can be deemed as a combination of these two models. Technically, RS-CNN continues to use the multi-scale encoder architecture and the relation module in MRNet, and merges information from each encoder after reasoning like MRNet does, while substitutes the sophisticated pattern module in MRNet with bottleneck residual convolutional blocks (the REM module). Meanwhile, the resemblance between RS-CNN and SAVIR-T lies in that both of them observe two rows simultaneously to obtain a rule representation, and this operation is natural since it is the only way to avoid ambiguity in deriving the rule representation.

\begin{figure}[h]%
\centering
\includegraphics[width=0.9\textwidth]{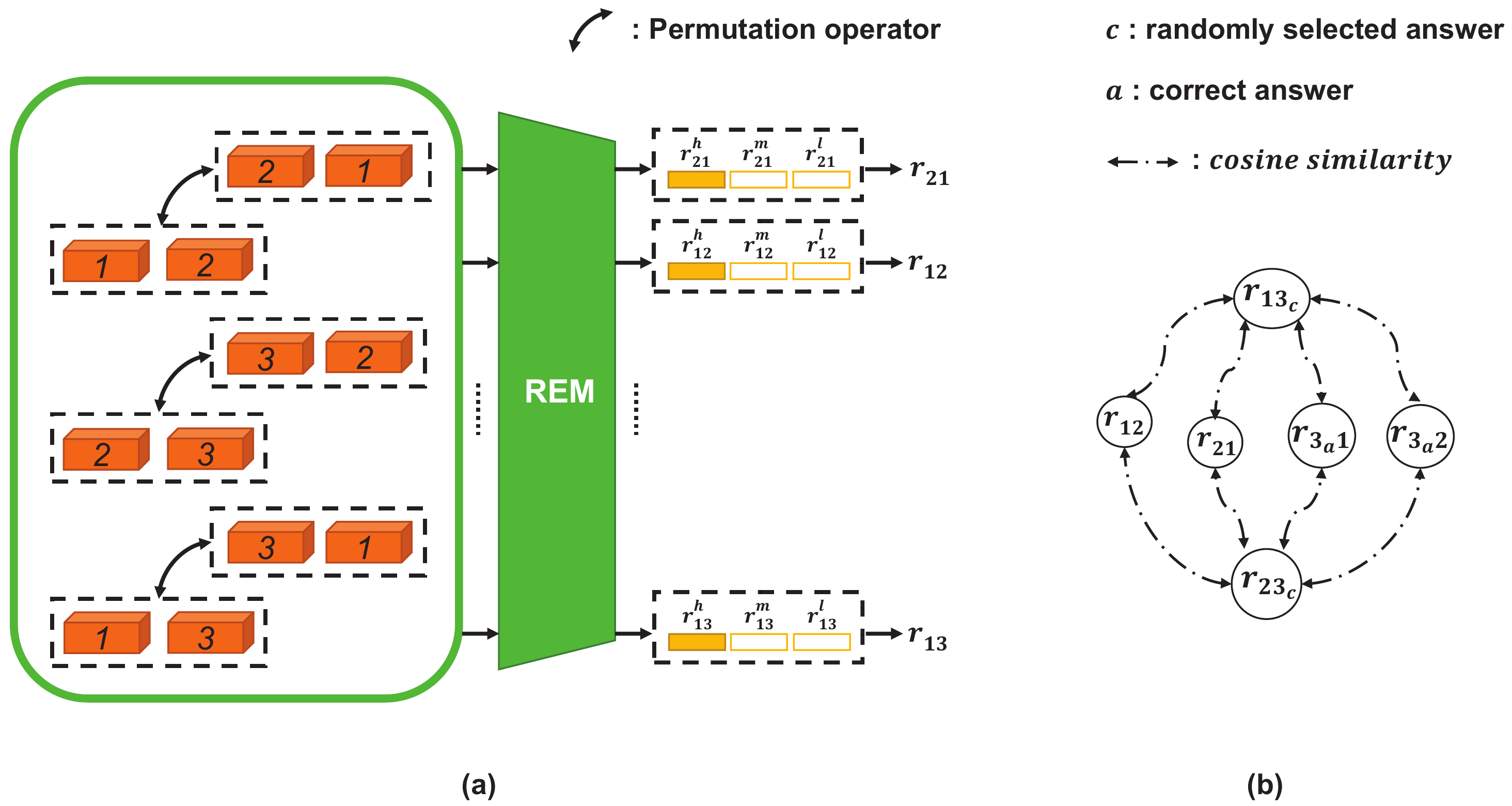}
\caption{(a) Illustration of the method of endowing RS-CNN with permutation invariance property. Aggregated information and the corresponding permuted ones are modeled explicitly and send to REM to obtain rule representations, and each aggregated information with its permuted counterpart are expected to output same rule representations. (b) Compute cosine similarity score between the guiding representations and two sets of rule representations $\{ {r_{13_c}}\} _{c = 1}^8$, $\{ {r_{23_c}}\} _{c = 1}^8$. }
\label{PermutationInvariance_CNN_Illustration}
\end{figure}

We wrap up the discussion of RS-CNN for RAVEN by introducing the inductive bias design, which conveys the property of permutation-invariance. Permutation-invariance indicates that exchanging the order of any two rows of the problem matrix does not effect the internal rules \cite{RPMInductivebias}. Previous works also devoted themselves to design neural networks which live up to permutation-invariance criterion \cite{CoPINet, MRNet, SAVIR-T}. Here we follow the spirit of modularization and endow the reasoning module with the permutation-invariance property. As shown in Fig. \ref{PermutationInvariance_CNN_Illustration}(a), we explicitly model the original aggregated information and the corresponding permuted aggregated information, and feed each of them into the REM module. Eq. \ref{PermutationNotation} defines notations for rule representations ${r_{ \cdot 3_*}}$, and ${r_{3_* \cdot }}$:

\begin{align}\label{PermutationNotation}
{r_{ \cdot 3_*}}, {r_{3_* \cdot }} = \left\{ \begin{array}{l}
{r_{ \cdot {3_a}}}, {r_{{3_a} \cdot }}\\
{r_{ \cdot {3_c}}}, {r_{{3_c} \cdot }},
\end{array} \right.
\end{align}
where $3_a$ or $3_c$ denotes the $3$rd row of the problem matrix with its last panel filled by the right answer or a randomly selected answer from the answer candidates, respectively. During the training phase, we set $r_{12}$, $r_{21}$, $r_{3_a1}$ and $r_{3_a2}$ as guiding representations, and compare each of them with two rule representation sets: $\{ {r_{13_c}}\} _{c = 1}^8$, and $\{ {r_{23_c}}\} _{c = 1}^8$, respectively, in terms of cosine similarity, as shown in Fig. \ref{PermutationInvariance_CNN_Illustration}(b). The training goal is to minimize the gap between the guiding representations and $r_{13_c}$, $r_{23_c}$ with the right answer (that is $r_{13_a}$, $r_{23_a}$, when $c=a$), while drift the guiding representations away from $r_{13_c}$ and $r_{23_c}$ with wrong answers (when $c\ne a$). We borrow the idea from contrastive learning \cite{SimCLR}, and consider each guiding representation as an independent query, and $\{ {r_{13_c}}\} _{c = 1}^8$, $\{ {r_{23_c}}\} _{c = 1}^8$ as two dictionaries, where the correct answer and wrong answers are treated as positive and negative samples. Each query and dictionary is paired up to compute an InfoNCE \cite{InfoNCE,MoCo} loss. All these symmetric losses are added up to obtain the loss for RS-CNN:

\begin{align}
{L_{RS - CNN}} = \sum\limits_{{r_g}} {[ - \sum\limits_{c = 1}^8 {{y_c}(\log \frac{{{e^{({r_g} \cdot {r_{{{13}_c}}})/\tau }}}}{{\sum\limits_i^8 {{e^{({r_g} \cdot {r_{{{13}_i}}})/\tau }}} }}} } ) - \sum\limits_{c = 1}^8 {{y_c}(\log \frac{{{e^{({r_g} \cdot {r_{{{23}_c}}})/\tau }}}}{{\sum\limits_i^8 {{e^{({r_g} \cdot {r_{{{23}_i}}})/\tau }}} }}} )],\\
\text{where} \quad {r_g} \in \{ {r_{12}},{r_{21}},{r_{{3_a}1}},{r_{{3_a}2}}\},\quad {y_c} = \left\{ \begin{array}{l}
0,c = a\\
1,c \ne a
\end{array} \right. \notag.
\end{align}
In this setting, the InfoNCE loss is actually the CrossEntropy loss augmented with a temperature parameter $\tau$ as a hyperparameter. During the test phase, the guiding representations are reduced to $r_{12}$, $r_{21}$, since the index of the right answer is unknown. Note that SAVIR-T is also trained by comparing the rule representations, however, the loss function in RS-CNN is more elaborated and flexible, with more rule representations available for comparison, and a flexible InfoNCE loss.

\begin{figure}[h]%
\centering
\includegraphics[width=0.9\textwidth]{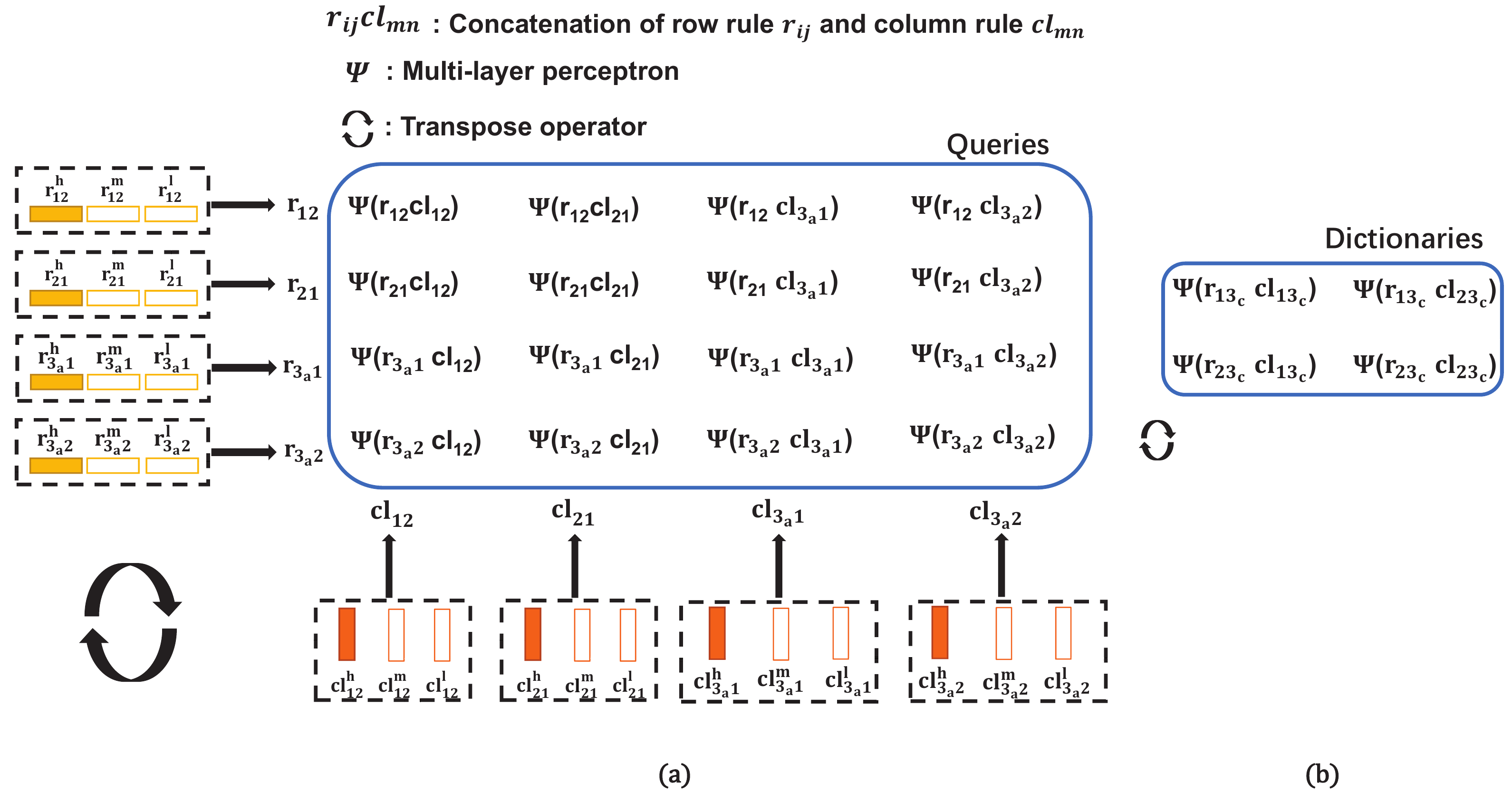}
\caption{Illustration of applying RS-CNN to PGM dataset. Concatenate each row rule representation with column rule representation, then obtain each row-column rule representation by feeding the concatenated vector to a two layer MLP $\Psi$. (a) List of row-column rule representations which are designated as guiding representations (queries). (b) Four sets of row-column rule representations (dictionaries), where each answer candidate in the answer pool is involved. Note that two concatenation orders (row-column and column-row) in forming row-column representations should be considered to live up with transpose invariance.}
\label{RS_CNN_PGM}
\end{figure}

We introduce the necessary revisions for adapting RS-CNN to the PGM dataset. Firstly assume that in PGM, both row-wise and column-wise rules always exist (which is not true). As shown in Fig. \ref{RS_CNN_PGM}, to make RS-CNN support PGM, we first follow the aforementioned logic to obtain the row rule representations, then we repeat the process column-wise to obtain the column rule representations. Next, pair up each row rule representation with the column one by concatenation to obtain a bunch of row-column rule representations. We designate guiding representations (queries) and dictionaries for contrastive learning, and the model is optimized to align each query with members in each dictionary with the right answer and drift away from others. Further more, notice that transpose the problem matrix, which is equivalent to change the concatenation order of the row-column representation, shall not effect the answer.

However, PGM is designed such that, the visual attributes of each panel in each row or column are allowed to change randomly, as long as there exists at least one rule. The aforementioned design is therefore not suitable for PGM, because this design neglects the `random change' phenomenon. Remedial efforts has been made to predict the `random change' pattern by using an extra neural network with row-column rule representation as input, and proves futile. In practice, the row-column rule representation is further processed with a shallow MLP layer, which is effective in mitigating the negative influence of the `random change'.

To sum up, the idea of explicitly design of transpose and permutation invariance as inductive bias in RS-CNN has pros and cons. Implementing these inductive bias in the reasoning module rather than shuffle the problem matrix directly saves computational resources by skipping repetitive computation in the perception module. However, although the idea behind RS-CNN is compatible with RAVEN and I-RAVEN, it contradicts with the nature of PGM, and the contradiction can only be alleviated to some extent. The drive to solve PGM and RAVEN completely gives birth to RS-TRAN.

\subsection{RS-TRAN}

Successful end-to-end RPM solvers in terms of reasoning accuracy (MRNet, SAVIR-T, SCL) have one thing in common: trying to encode image representations in different perspectives. Specifically, MRNet adopts multi-scale encoders, SAVIR-T combines residual convolutional block with Transformer block to produce multiple visual tokens, while SCL scatters extracted visual representations into pieces. The present RS-TRAN follows the spirit of these methods, with a more scalable implementation. The notations in RS-CNN and RS-TRAN are independent of each other.

\begin{figure}[h]%
\centering
\includegraphics[width=0.9\textwidth]{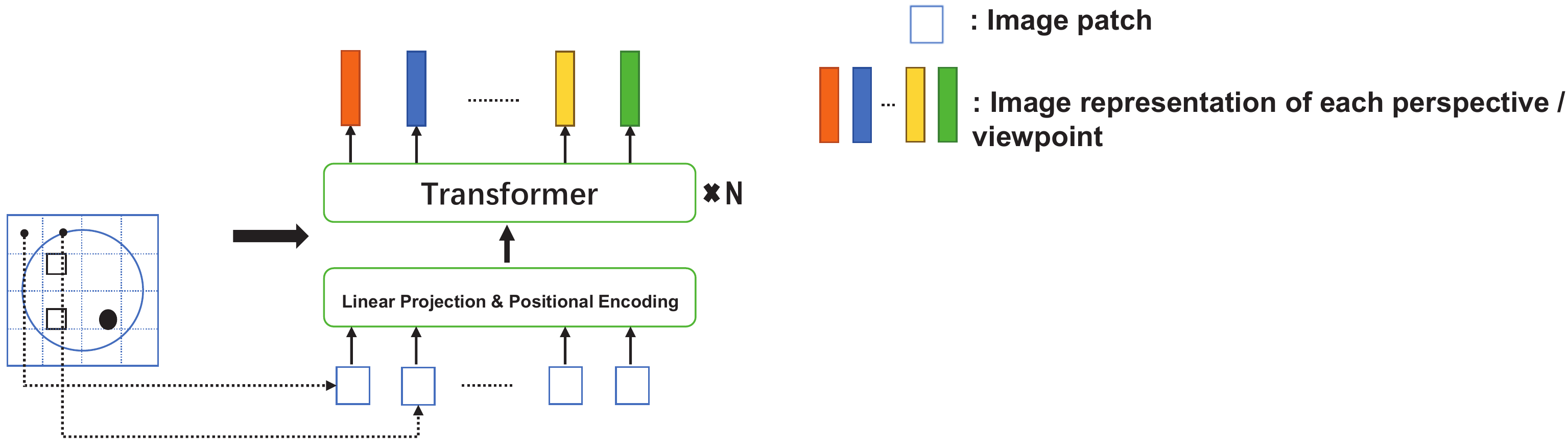}
\caption{Illustration of the perception module in RS-TRAN. Each image in the problem matrix is processed independently: split into patches, processed by the same linear embedding layer with positional encoding, then several Transformer blocks. The output of each image will be a bunch of feature representations, each of which stands for one viewpoint.}
\label{RS_TRAN_Perception}
\end{figure}

The perception module of RS-TRAN is developed with the backbone of ViT and the implementation of multi-viewpoint mechanism. As shown in Fig. \ref{RS_TRAN_Perception}, each image in the problem matrix and the answer candidates is first split into 16 patches and then fed into a linear embedding layer with positional encoding, whose output will be sent to Transformer blocks. In the output side of the last Transformer block, between using global average pooling and special classification token \cite{Bert,ViT}, we choose neither of them and leave all the outputs unprocessed, and end up with multiple outputs for each image. Such move can be interpreted as maintaining the diversity of the global features in the perspective of every patch, which is the realization of the multi-viewpoint mechanism. The number of viewpoint equals the number of patches for each image, which is 16 in RS-TRAN.

\begin{figure}[h]%
\centering
\includegraphics[width=0.9\textwidth]{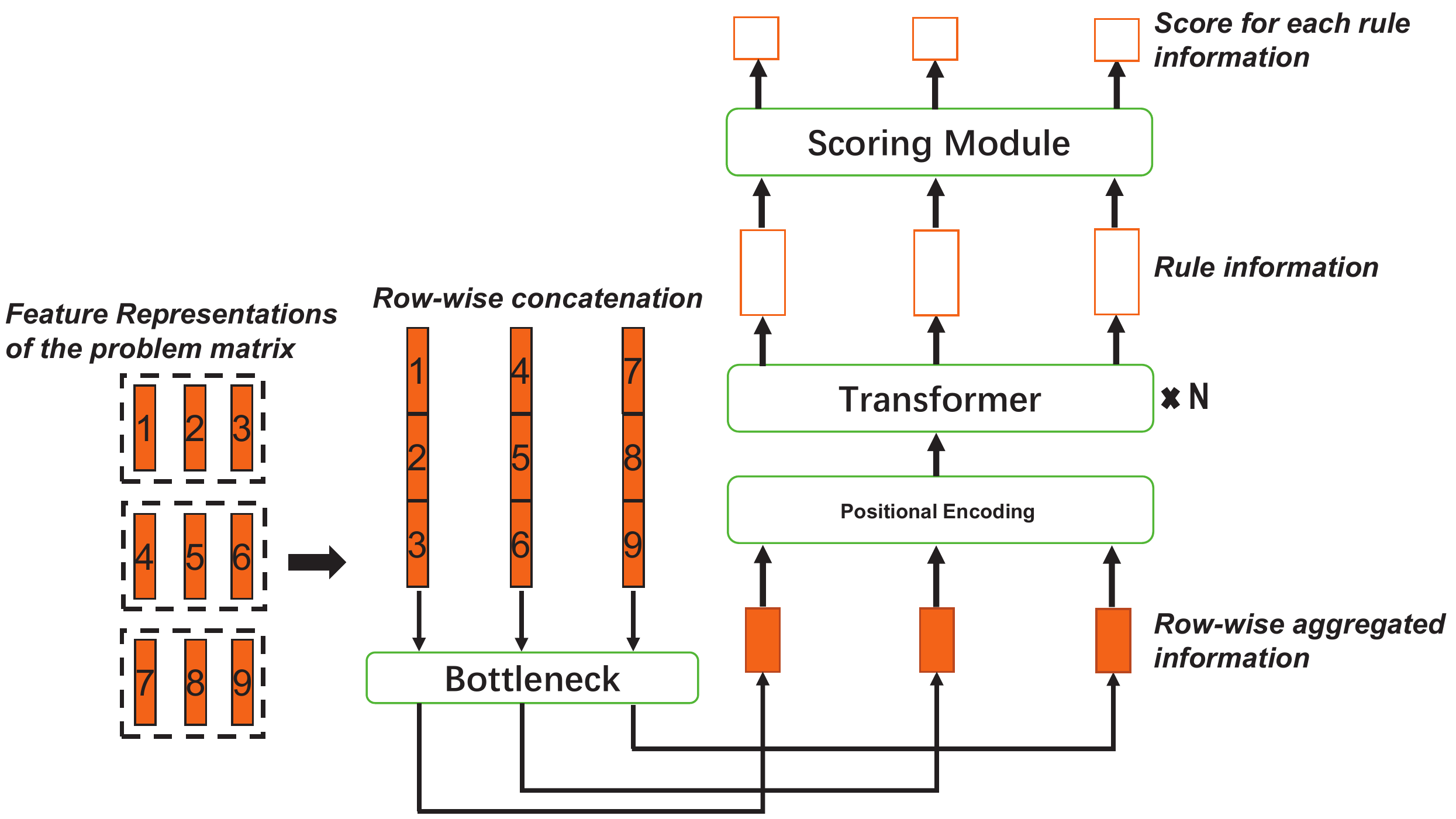}
\caption{Illustration of one viewpoint of the reasoning module in RS-TRAN. In each viewpoint, the problem matrix with each answer candidate is evaluated multiple times to obtain multiple scores. The bottleneck architecture is introduced to replace the linear embedding layer of the vanilla transformer.}
\label{RS_TRAN_Reasoning}
\end{figure}

In the reasoning module of RS-TRAN, we process feature representations of each viewpoint individually. when dealing with RAVEN or I-RAVEN dataset, set the problem matrix with the right answer or wrong answer from the answer candidates as positive and negative sample, respectively. As shown in Fig. \ref{RS_TRAN_Reasoning}, for each viewpoint, representations are concatenated row-wise and fed into a bottleneck module to produce three aggregated information, which will be sent to Transformer blocks after positional encoding. Similar to the perception module, the outputs of the last transformer block will be left unprocessed, and each of them will be sent to a scoring module formed by bottleneck layers to obtain a score, which is the realization of multi-evaluation. That means, for each problem matrix with an answer, the amount of scores equals the product of number of viewpoints in the perception module and number of scores in each reasoning module. All these scores are averaged to obtain a final score. Setting CrossEntropy loss as the loss function, the network is optimized so that the score of the problem matrix with the right answer outnumbers others: for the problem matrix $\left\{ {X_{_{p}}^i} \right\}_{i = 1}^8$ and the corresponding answer candidates $\left\{ {X_{_{ac}}^i} \right\}_{i = 1}^8$, the score is calculated as follows:

\begin{align}
{s_j} = \frac{1}{{3 \times 16}}\sum\limits_{m = 1}^{16} {\sum\limits_{n = 1}^3 {s_j^{m,n}} } ,
\end{align}
where $m$ indicates the index of viewpoint, $n$ indicates the index of score in each viewpoint, and $j$ indicates the index of answer in the answer candidate, $s_j^{m,n}$ denotes the $n$th score in $m$th viewpoint for the problem matrix with $j$th answer, while $s_j$ denotes the final score of the $j$th answer. Denote $a$ the right answer, $c$ a randomly selected answer, from the answer candidates. The loss function of RS-TRAN is defined as:

\begin{align}
 L_{RS-TRAN}=- \sum\limits_{c = 1}^8 {{y_c}\log {s_c}},\\
\text{where} \quad {y_c} = \left\{ \begin{array}{l}
1,c = a\\
0,c \ne a
\end{array} \right.\notag
\end{align}

\begin{figure}[h]%
\centering
\includegraphics[width=0.9\textwidth]{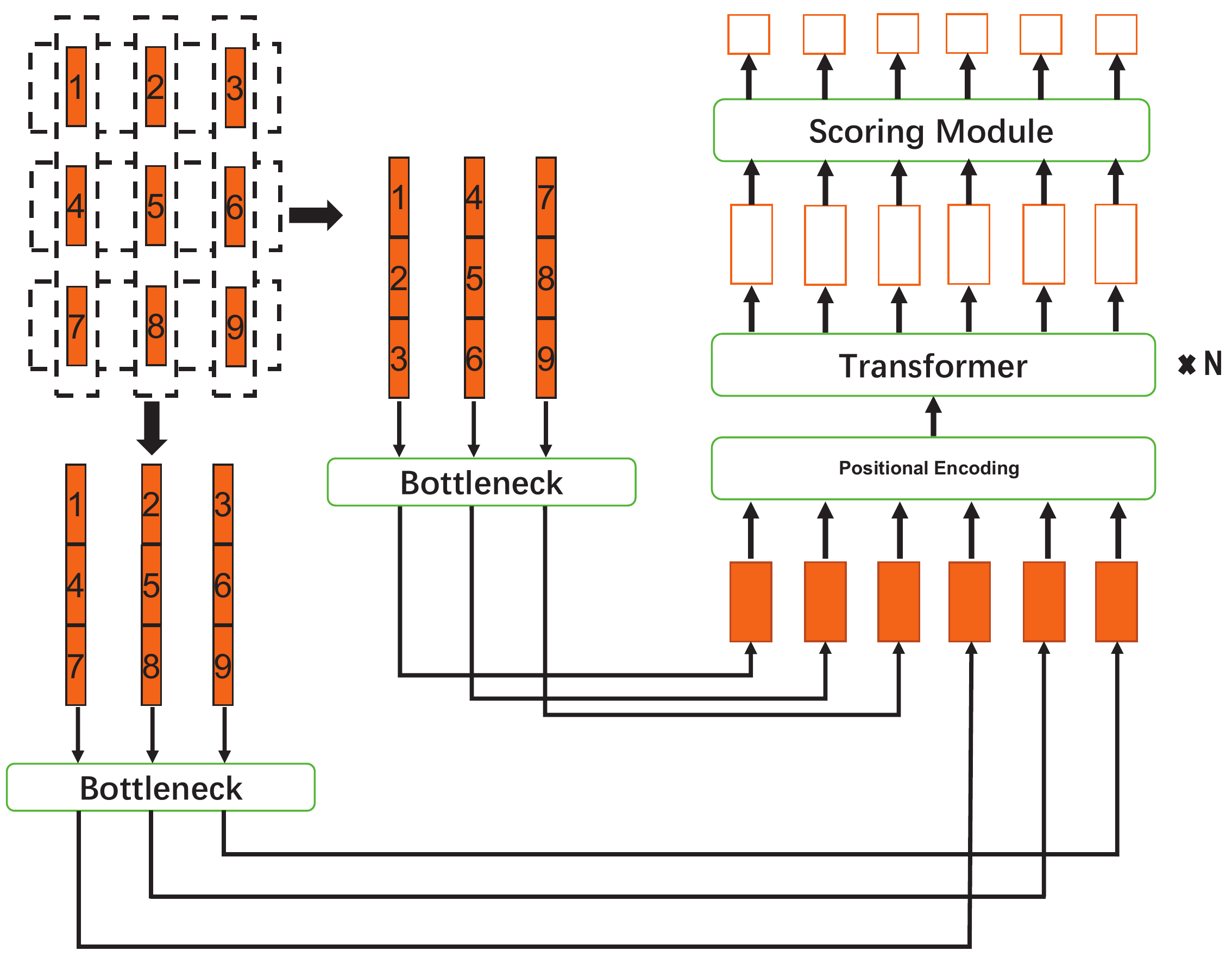}
\caption{Illustration of applying RS-TRAN to PGM dataset. Only one viewpoint of the image representations is shown. The modification made here is to obtain column aggregated information, and then blend it with the row aggregated information.}
\label{RS_TRAN_PGM}
\end{figure}

The necessary revisions for adapting RS-TRAN to the PGM dataset is straightforward. Fig. \ref{RS_TRAN_PGM} shows how to process one viewpoint of feature representations in the reasoning module. Intuitively, PGM needs both aggregated row and column information to complete the reasoning process. All the aggregated information is sent to the Transformer blocks, while other operations remain the same as applying RS-TRAN to solve the RAVEN or I-RAVEN dataset.

The explicit implementation of inductive bias for expressing permutation invariance or transpose invariance in RS-TRAN is not necessary: changing the order of the aggregated information is equivalent to the capability of permutation or transpose operator, and in RS-TRAN, the order of the aggregated information only influences the order of output scores (leave alone the effect of positional encoding), which is of no effect after the averaging operation. In this scenario, learnable positional encoding \cite{PositionalEncoding} is preferred.

RS-TRAN bears a resemblance to RS-CNN in that both of them consist of a perception module and a reasoning module. However, the implicit implementation of inductive bias leads RS-TRAN to a refreshing and uncomplicated network architecture.

\subsection{RS-TRAN-CLIP}

\begin{figure}[h]%
\centering
\includegraphics[width=0.7\textwidth]{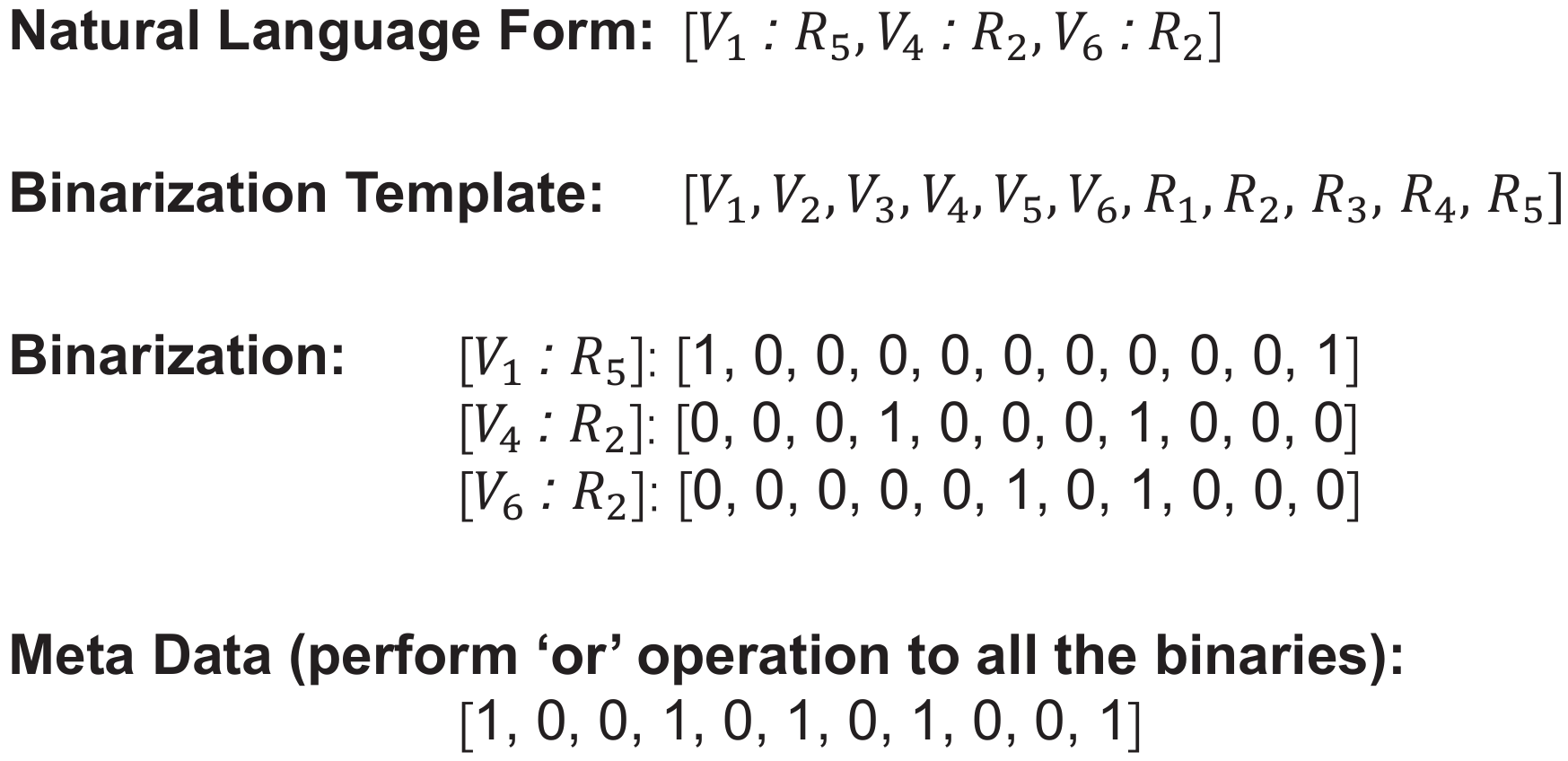}
\caption{Production process of meta-data: perform `or' operation to all of the binarized [visual attribute: rule] pair. }
\label{MetaDataDemo}
\end{figure}

The original RAVEN, I-RAVEN and PGM datasets come with meta-data, which contains the information of rules of the problem matrix \cite{RAVENdataset,PGMdataset}. For example, denote visual attributes as: $V_1, V_2, ..., V_6$, and rules as: $R_1, R_2,..., R_5$, Fig. \ref{MetaDataDemo} shows a simple demonstration of the meta-data.

Previous end-to-end models usually utilize the meta-data by developing an auxiliary branch to predict them. The results reported in these works show that, the auxiliary task of predicting meta-data improves the performance of reasoning, when the baseline reasoning accuracy is not very high  \cite{RAVENdataset,PGMdataset,PrAE}; however the very same auxiliary task shows negative effect, when the baseline reasoning accuracy reaches certain level \cite{MRNet}. Dive into the structure of meta-data, we notice that, the meta-data is originally designed to improve the performance of models, however, its binary form carries less information than its natural language form. As shown in Fig. \ref{MetaDataDemo}, by implementing `or' operation to all of the binarized [visual attribute: rule] pair, we lose track of the exact paring information between the visual attributes and the rules, which causes unnecessary optimization trouble for the model. In view of this, we abandon the binarized mata-data, and develop RS-TRAN-CLIP to predict the meta-data in its natural language form.

\begin{figure}[h]%
\centering
\includegraphics[width=1.0\textwidth]{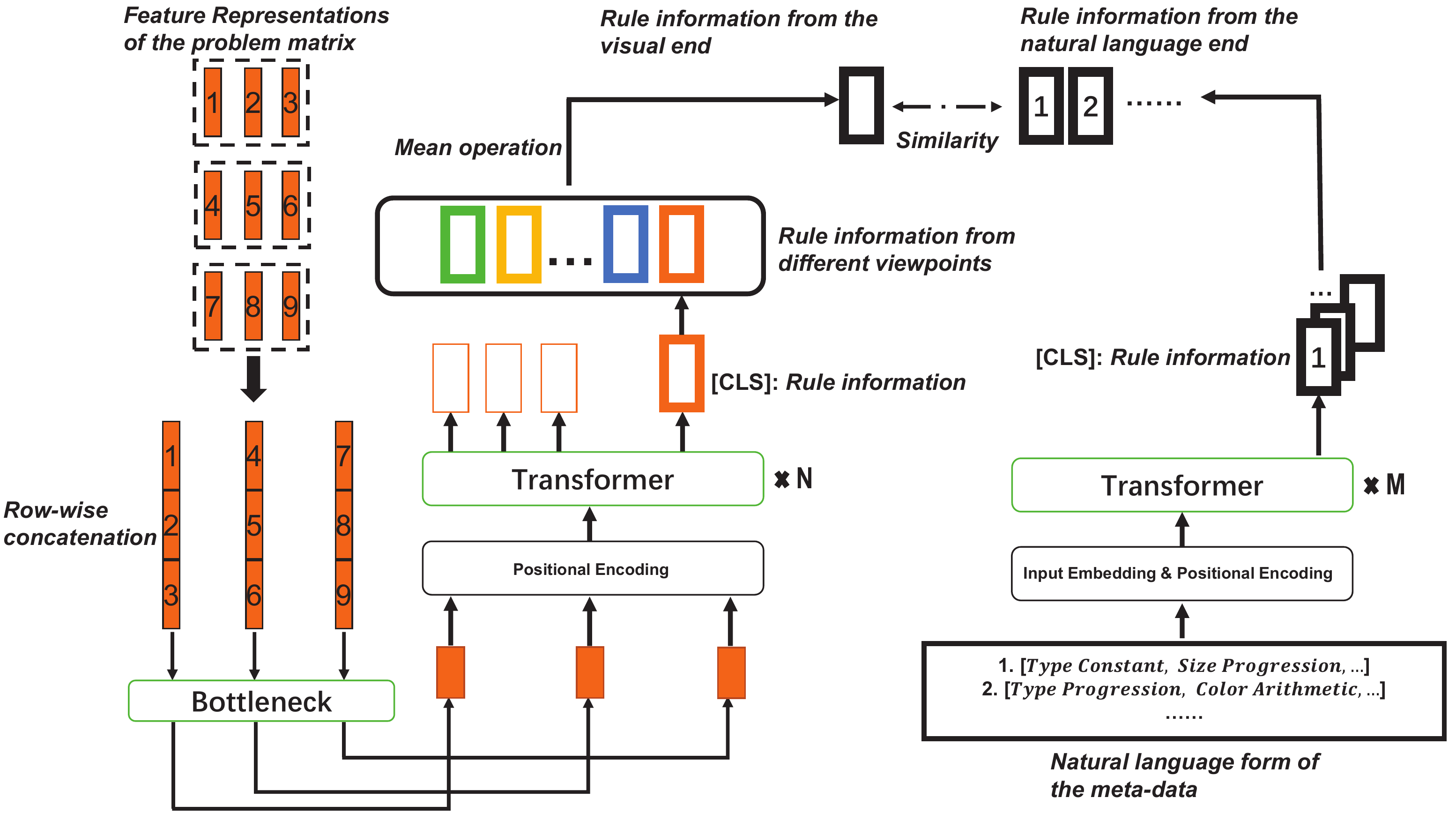}
\caption{The architecture of the RS-TRAN-CLIP in one viewpoint (the perception module in the visual end is not shown).}
\label{RS-TRAN-CLIP}
\end{figure}

The core structure of RS-TRAN-CLIP resembles CLIP \cite{Clip}. RS-TRAN-CLIP has three homologous architectures specializing in RAVEN and I-RAVEN with one set of rule (e.g. `3 $\times$ 3 Grid'), RAVEN and I-RAVEN with two sets of rule (e.g., `O-IG'), and PGM, respectively. We introduce these architectures one by one.

\textbf{I}. RS-TRAN-CLIP for RAVEN and I-RAVEN with one set of rule. RS-TRAN-CLIP has a visual end and a natural language end. We introduce the visual end first. The visual end takes completed problem matrix with the correct answer as input. The perception part of the visual end inherits from RS-TRAN, which outputs image representations in multiple viewpoints. Fig. \ref{RS-TRAN-CLIP} shows both the reasoning part of the visual end and the natural language end. In the reasoning part, for image representations from each viewpoint, instead of applying the multi-evaluation mechanism, a `CLS' vector is produced, then the average of the `CLS' vectors from all viewpoints is regarded as the rule representation of the concerning problem matrix. While the natural language end takes in all the rules in the dataset as inputs, then produces a representation for each rule, using standard Transformer with `CLS' vector. The model is optimized so that the rule representation from the visual end aligns itself with the corresponding rule representation from the natural language end, in terms of cosine similarity. We again use the InfoNCE as the loss function.

\begin{figure}[h]%
\centering
\includegraphics[width=0.4\textwidth]{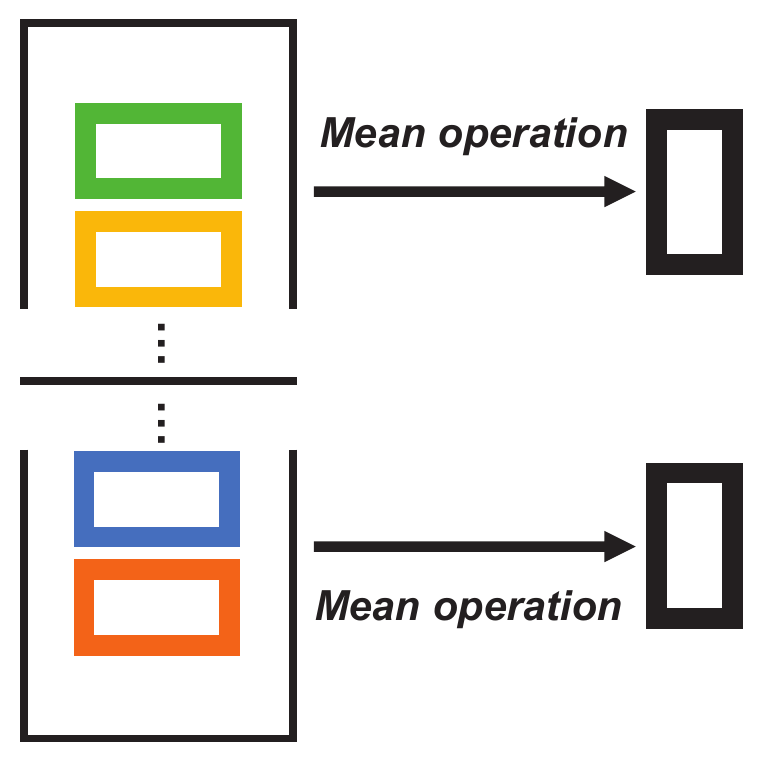}
\caption{Divide the `CLS' vectors of all viewpoints into two groups, so that the RS-TRAN-CLIP model can deal with configurations with two sets of rules, such as 'O-IG' in Raven and I-RAVEN datasets.}
\label{RS-TRAN-CLIP-Two-Rules}
\end{figure}

\textbf{II}. RS-TRAN-CLIP for RAVEN and I-RAVEN with two sets of rule. Adjustments are made so that the visual end can produce two rule representations. As shown in Fig. \ref{RS-TRAN-CLIP-Two-Rules}, the `CLS' vectors of all viewpoints are evenly divided into two groups, with each group corresponds to one set of rule.

\textbf{III}. The reasoning module also requires adjustment when applying RS-TRAN-CLIP for PGM. First, both row and column information are needed when dealing with PGM, detailed adjustment for obtaining row and column information simultaneously is shown in Fig. \ref{RS_TRAN_PGM}. Second, the `line' and `shape' in PGM are independent of each other. As a result of that, the reasoning module also produces two rule representations, one for `line', the other for `shape'. For some cases where `line' or `shape' changes randomly, the corresponding rule is set as `NA'.

\begin{figure}[h]%
\centering
\includegraphics[width=0.9\textwidth]{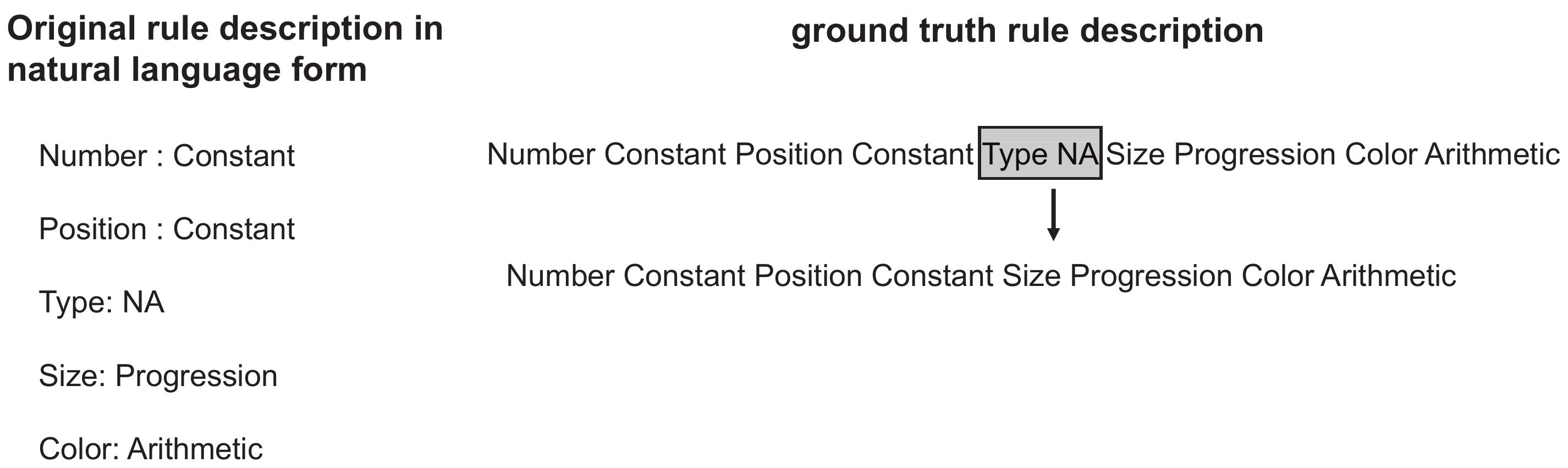}
\caption{Demonstration of the mask scheme in the RS-TRAN-CLIP. By masking out all visual attributes with `NA', not only we buffer the model against extra optimization burden, but also obtain compatible image representations for downstream tasks.}
\label{NA_Mask}
\end{figure}

Finally, we discuss a special practice in RS-TRAN-CLIP. PGM dataset has `NA' rule as aforementioned. While the RAVEN and I-RAVEN dataset have `NA' rule in their original package. `NA' indicates the random change of certain visual attributes, and any efforts in predicting chaotic rule representations as a definite `NA' is of no help in improving model's reasoning capability. In view of this, we mask all the `NA' term in the training data, which means that RS-TRAN-CLIP only needs to correctly predict `non-NA' rules, as shown in Fig. \ref{NA_Mask}.

The trained RS-TRAN-CLIP serves as a pre-training model for RS-TRAN. Specifically, parameters of the perception module in the trained RS-TRAN-CLIP are imported into the RS-TRAN. Then the RS-TRAN module is trained with this new initialization. Note that the parameters in the perception module is not frozen when training the RS-TRAN.

\section{Experiments}\label{experiments}
In this section, we report the reasoning accuracy of the proposed models in solving RPM problems, and compare these results with former state-of-the-arts: SCL \cite{SCL}, SAVIR-T \cite{SAVIR-T}, and MRNet \cite{MRNet}. For ablation studies, we evaluate the effect of dataset size, inductive bias, and multi-viewpoint with multi-evaluation scheme. We also conduct a series of intuitive tests to study the intrinsic behaviours of our models, and test our model's generalization ability.

\subsection{Reasoning Accuracy}
Both RS-CNN and RS-TRAN are implemented in Pytorch\cite{Pytorch}, and optimized by ADAM\cite{ADAM}. The training session is terminated when the reasoning accuracy of the validation set ceases to improve, and we report the reasoning accuracy in the test set.

For each configuration in RAVEN and I-RAVEN, data is split into training set, validation set and test set, with the size of 80K / 20K / 40K for RS-CNN and 155K / 5K / 40K  for RS-TRAN.

\begin{table}[h]
\caption{Reasoning Accuracies on RAVEN and I-RAVEN.}
\label{RAVEN_IRAVEN_Results}
\centering
\resizebox{\linewidth}{!}{
\begin{tabular}{cccccccccc}
\toprule
&\multicolumn{8}{c}{Test Accuracy(\%)}& \\
\cmidrule{2-9}
Model&Average&Center&2 $\times$ 2 Grid&3 $\times$ 3 Grid&L-R&U-D&O-IC&O-IG \\
\midrule
SAVIR-T \cite{SAVIR-T}&94.0/98.1&97.8/99.5&94.7/98.1&83.8/93.8&97.8/99.6&98.2/99.1&97.6/99.5&88.0/97.2\\
\midrule
SCL \cite{SCL, SAVIR-T}&91.6/95.0&98.1/99.0&91.0/96.2&82.5/89.5&96.8/97.9&96.5/97.1&96.0/97.6&80.1/87.7\\
\midrule
MRNet \cite{MRNet}&96.6/-&-/-&-/-&-/-&-/-&-/-&-/-&-/-\\
\midrule
RS-CNN&95.1/\textbf{99.3}&\textbf{100.0}/\textbf{100.0}&94.2/97.9&84.5/\textbf{97.9}&\textbf{100.0}/\textbf{100.0}&\textbf{99.4}/\textbf{100.0}&99.7/\textbf{100.0}&87.6/\textbf{99.2} \\
\midrule
RS-TRAN&\textbf{98.4}/98.7&99.8/\textbf{100.0}&\textbf{99.7}/\textbf{99.3}&\textbf{95.4}/96.7&99.2/\textbf{100.0}&\textbf{99.4}/99.7&\textbf{99.9}/99.9&\textbf{95.4}/95.4 \\
\bottomrule
\end{tabular}
}
\end{table}

Tab. \ref{RAVEN_IRAVEN_Results} shows the reasoning accuracies of RS-CNN, RS-TRAN, and former state-of-the-art models. We must point out that the direct comparison between these models are quiet unfair: the training set size of RS-CNN and RS-TRAN is way more bigger than MRNet and SCL. We can only conclude that, both RS-CNN and RS-TRAN can achieve state-of-the-art results, with abundant training data.

For PGM, we set the size of training set, validation set and test set as 1200K / 20K / 400K, which is exactly the same as other models do. As shown in Tab. \ref{PGM_Results}, RS-TRAN surpasses the former SOTA results, while RS-CNN only achieves the reasoning accuracy of 82.8\%, which is foreseeable because the inductive bias in RS-CNN is not compatible with PGM. Note that PGM is a very large dataset, for the convenience of parameter-tuning, in each epoch, we only train a randomly drawn mini-batch of the training data.

\begin{table}[h]
\caption{Reasoning Accuracies on PGM.}
\label{PGM_Results}
\centering
\begin{tabular}{ccc}
\toprule
Model&Test Accuracy(\%) \\
\midrule
SAVIR-T \cite{SAVIR-T}&91.2\\
\midrule
SCL \cite{SCL, SAVIR-T}&88.9\\
\midrule
MRNet \cite{MRNet}&94.5\\
\midrule
RS-CNN&82.8\\
\midrule
RS-TRAN&\textbf{97.5}\\
\bottomrule
\end{tabular}
\end{table}

\begin{figure}[h]%
\centering
\includegraphics[width=0.9\textwidth]{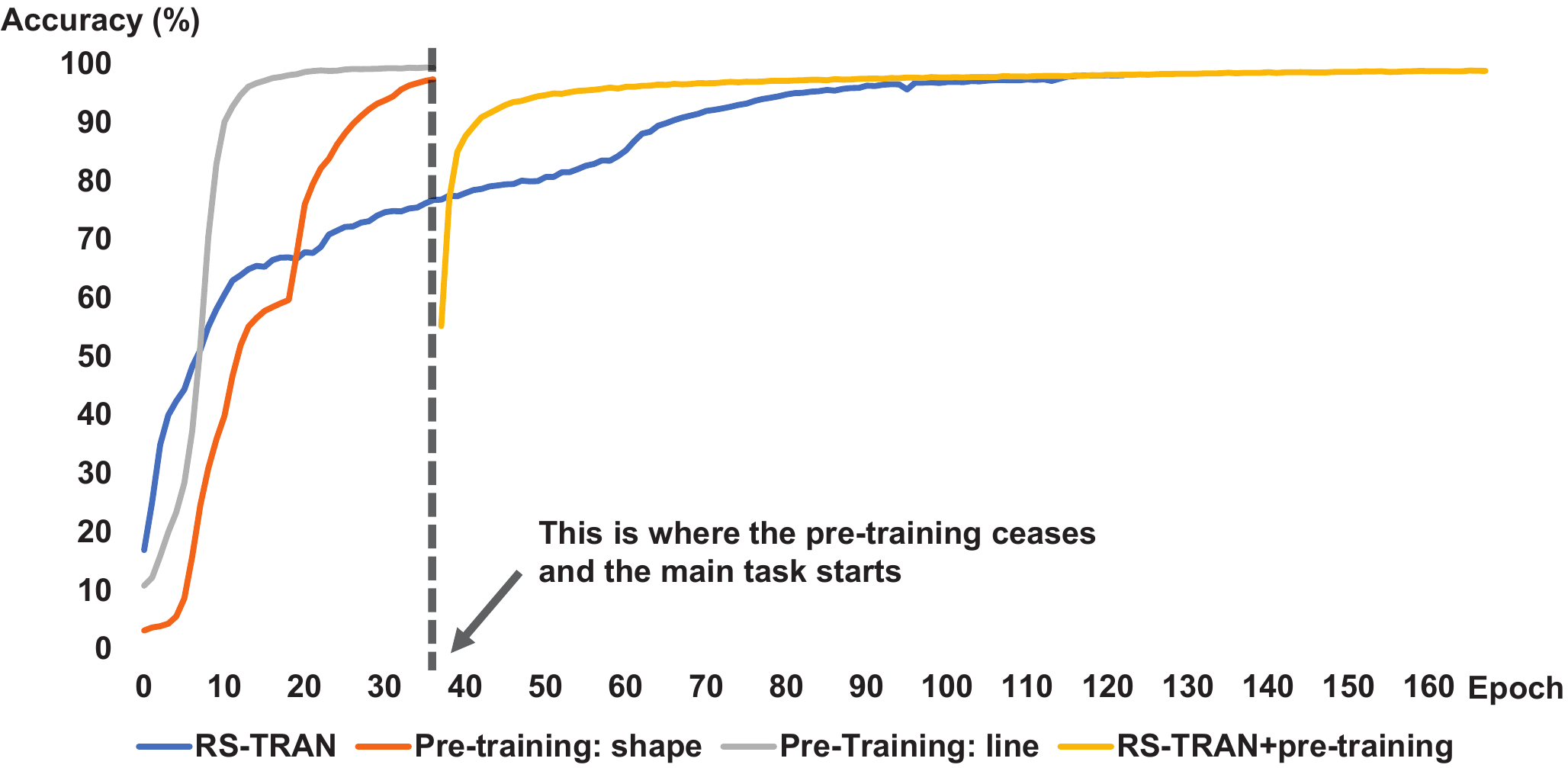}
\caption{Training trajectories of RS-TRAN for PGM, with \& without pre-training.}
\label{TRAN_CLIP_Training_Data}
\end{figure}

Although RS-CNN and RS-TRAN already achieve satisfying results, it is shown in Tab. \ref{TRAN_CLIP_Results} that, RS-TRAN with a pre-training perception module powered by RS-TRAN-CLIP achieves even higher reasoning accuries in the most difficult RPM tasks. Fig. \ref{TRAN_CLIP_Training_Data} also shows that, RS-TRAN with pre-training converges to its upper limit smoothly and fast.

\begin{table}[h]
\caption{Reasoning Accuracies of the RS-TRAN with pre-training, on part of I-RAVEN and PGM.}
\label{TRAN_CLIP_Results}
\centering
\begin{tabular}{cccccccccc}
\toprule
&\multicolumn{3}{c}{Test Accuracy(\%)}& \\
\cmidrule{2-4}
Model&3 $\times$ 3 Grid&O-IG&PGM \\
\midrule
RS-TRAN without pre-training&96.7&95.37&97.5 \\
\midrule
RS-TRAN with pre-training&\textbf{99.2}&\textbf{99.1}&\textbf{99.0} \\
\bottomrule
\end{tabular}
\end{table}

\subsection{Effect of dataset size}
In this section, we investigate the impact of the dataset size. The performance of neural networks depends on the size of dataset heavily\cite{DatasetSize1}, so is the case with Transformer-based networks \cite{DatasetSize2,ViT}. For simplicity, we only take `Center', `3 $\times$ 3 Grid' and `O-IG' in RAVEN and I-RAVEN for example.

The performance of RS-CNN in RAVEN is mediocre compared with its results in I-RAVEN, so we increase the size of training set in RAVEN to 155K/configuration. Also notice that RS-TRAN requires tons of training data, so we decrease the training set size of I-RAVEN to 80K/configuration.

\begin{table}[h]
\caption{Effect of dataset size on RS-CNN.}
\label{RSCNN_Datasize}
\centering
\begin{tabular}{cccccccccc}
\toprule
&\multicolumn{8}{c}{Test Accuracy(\%) of RAVEN}& \\
\cmidrule{2-9}
Model&Center&3 $\times$ 3 Grid&O-IG \\
\midrule
RS-CNN(80K)&100&84.5&87.6 \\
\midrule
RS-CNN(155K)&100&94.2&99.2 \\
\bottomrule
\end{tabular}
\end{table}

\begin{table}[h]
\caption{Effect of dataset size on RS-TRAN.}
\label{RSTRAN_Datasize}
\centering
\begin{tabular}{cccccccccc}
\toprule
&\multicolumn{8}{c}{Test Accuracy(\%) of I-RAVEN}& \\
\cmidrule{2-9}
Model&Center&3 $\times$ 3 Grid&O-IG \\
\midrule
RS-TRAN(80K)&99.6&94.5&95.2 \\
\midrule
RS-TRAN(155K)&100&96.7&95.4 \\
\bottomrule
\end{tabular}
\end{table}

The general conclusion is not surprising: as shown in Tab. \ref{RSCNN_Datasize} and Tab. \ref{RSTRAN_Datasize}, bigger dataset size leads to better performance. Specifically, RS-CNN improves dramatically when the size of training set increases from 80K to 155K, while RS-TRANS already has good performance when the dataset size is 80K, and the results can be further improved by expanding the training dataset size.

\subsection{Ablation Studies}
In this part, we dive deeper into the architecture of RS-CNN and RS-TRAN, discuss the necessity of injecting inductive biases into RS-CNN, and the plausibility of RS-TRAN on its `natural expressiveness' of these inductive biases. We also study the effectiveness of multi-viewpoint and multi-evaluation of RS-TRAN. All the ablation studies are conducted on PGM and certain configurations of I-RAVEN.

\begin{table}[h]
\caption{Effect of inductive bias on RS-CNN(80K) and RS-TRAN(155K).}
\label{Inductive_Bias_Results}
\centering
\begin{tabular}{cccccccccc}
\toprule
&\multicolumn{8}{c}{Test Accuracy(\%)} of I-RAVEN and PGM& \\
\cmidrule{2-9}
Model&Center&3 $\times$ 3 Grid&O-IG&PGM \\
\midrule
RS-CNN&100&97.9&99.2&82.8 \\
\midrule
RS-CNN($-$)\footnotemark[1]&100&74.7&75.7&76.1 \\
\midrule
RS-TRAN&100.0&96.7&95.4&97.5 \\
\midrule
RS-TRAN($+$)\footnotemark[2]&99.9&94.6&97.1&96.1 \\
\bottomrule
\end{tabular}
\footnotetext[1]{RS-CNN without the consideration of permutation and transpose invariance.}
\footnotetext[2]{RS-TRAN with explicit augmentation of permutation and transpose invariance.}
\end{table}

In RS-CNN, the elimination of designed inductive biases indicates the prohibition of permutation and transpose operators. In RS-TRAN, taking permutation and transpose invariance into consideration means the explicit traversal of input ordering in the reasoning module. The results of Tab. \ref{Inductive_Bias_Results} conveys strong signals: explicit inductive design is crucial in RS-CNN, while the architecture of RS-TRAN naturally enjoys the property of permutation and transpose invariance, however explicit inductive bias design will do no harm to the performance.

In the other hand, RS-TRAN without the mechanism of multi-viewpoint and multi-evaluation only has a single, averaged output in both the perception and the reasoning module. Tab. \ref{Single_View_Evaluation_Results} tells us that the reasoning accuracy of RS-TRAN will drop sharply in such a situation. The effectiveness of multi-viewpoint and multi-evaluation mechanism is then self-evident.

\begin{table}[h]
\caption{Effect of multi-viewpoint and multi-evaluation on RS-TRAN(155K).}
\label{Single_View_Evaluation_Results}
\centering
\begin{tabular}{cccccccccc}
\toprule
&\multicolumn{8}{c}{Test Accuracy(\%)} of I-RAVEN and PGM& \\
\cmidrule{2-9}
Model&Center&3 $\times$ 3 Grid&O-IG&PGM \\
\midrule
RS-TRAN&100.0&96.7&95.4&97.5 \\
\midrule
RS-TRAN($-$)\footnotemark[1]&99.7&73.0&48.4&45.7 \\
\bottomrule
\end{tabular}
\footnotetext[1]{RS-TRAN with single output in the output layer of the perception and reasoning module. The original outputs of these two modules are averaged, respectively.}
\end{table}

\subsection{More on Multi-viewpoint and Multi-evaluation}
Both the reasoning accuracy and the ablation study show the necessity of adopting multi-viewpoint with multi-evaluation mechanism in RS-TRAN, however the detailed effects of this mechanism is still opaque. To have a clearer understanding of it, we design mask experiments.

In RS-TRAN, multi-viewpoint mechanism is implemented by retaining all the outputs of the perception module. Each image of the problem matrix is divided into 16 patches, which leads to 16 viewpoints (outputs) in the perception module. functionality of each output can be roughly probed by masking it out and testing the reasoning accuracy of RPM problem with specific rules. The same procedures are applicable to multi-evaluation mechanism, too.

\begin{figure}[h]%
\centering
\includegraphics[width=0.6\textwidth]{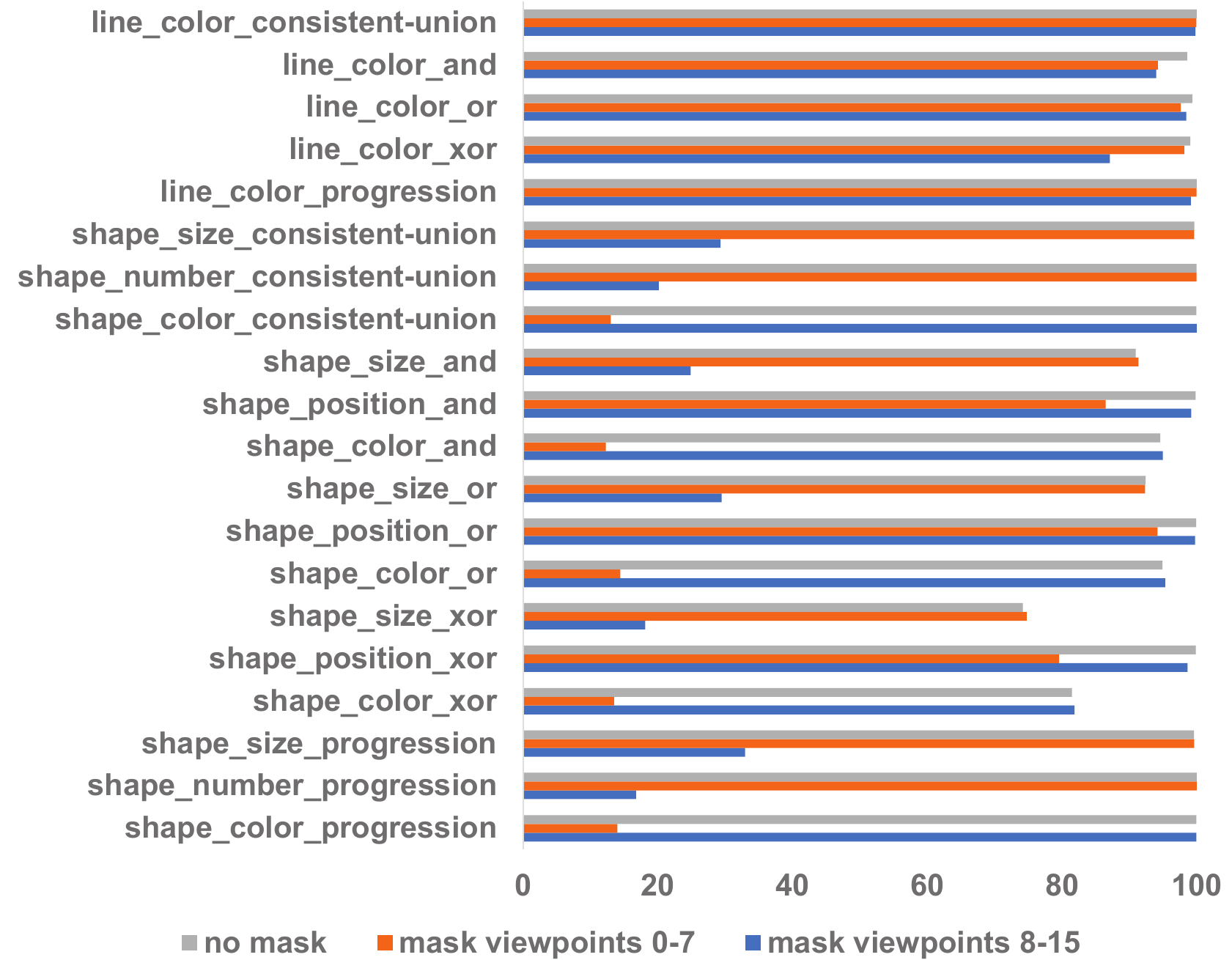}
\caption{Mask experiments on single-rule data of PGM. Only problems with one rule will be considered to avoid semantic ambiguity.}
\label{Experiment_Multi_Multi}
\end{figure}

For simplicity, we only study the multi-viewpoint mechanism on single-rule problems from PGM, by masking out half of the outputs in the perception module of RS-TRAN. As shown in Fig. \ref{Experiment_Multi_Multi}, Viewpoints $[0-7]$ and $[8-15]$ tend to focus on different rules. With the masking technique, the model acquires certain level of post-hoc interpretability \cite{MaskInterpretability1,MaskInterpretability2}. And we conclude that the multi-viewpoint with multi-evaluation mechanism endows the model with the ability of solving problems in several perspectives simultaneously.

\subsection{On Generalization}

In PGM, apart from the `neutral' dataset, where the data from the training set and test set is independent and identically distributed (I.I.D.), there are other datasets whose distribution of training sets and test sets are different (Out of distribution, O.O.D.). We evaluate the generalization ability of the RS-TRAN model by training on these datasets. The experiment setting remains the same with the `neutral' case.

\begin{table}[h]
\caption{Generalization Results of RS-TRAN in PGM.}
\label{Generalization_PGM}
\centering
\begin{tabular}{cccccccccc}
\toprule
Dataset&Accuracy(\%) \\
\midrule
Interpolation & 77.2 \\
\midrule
Extrapolation & 19.2 \\
\midrule
Held-out Attribute shape-colour & 12.9 \\
\midrule
Held-out Attribute line-type & 24.7 \\
\midrule
Held-out Triples & 22.2 \\
\midrule
Held-out Pairs of Triples & 43.6 \\
\midrule
Held-out Attribute Pairs & 28.4 \\
\bottomrule
\end{tabular}
\end{table}

The results of Tab. \ref{Generalization_PGM} declare the vulnerability of RS-TRAN facing the generalization problems. Facing dataset with sophisticated compositional structures, well-designed rules and no trace of noise in pixel-level, attempts without remarkably high reasoning accuracies should be deemed as a failure.

\subsection{Does the model learn the pre-defined rules?}
\begin{figure}[H]%
\centering
\includegraphics[width=0.6\textwidth]{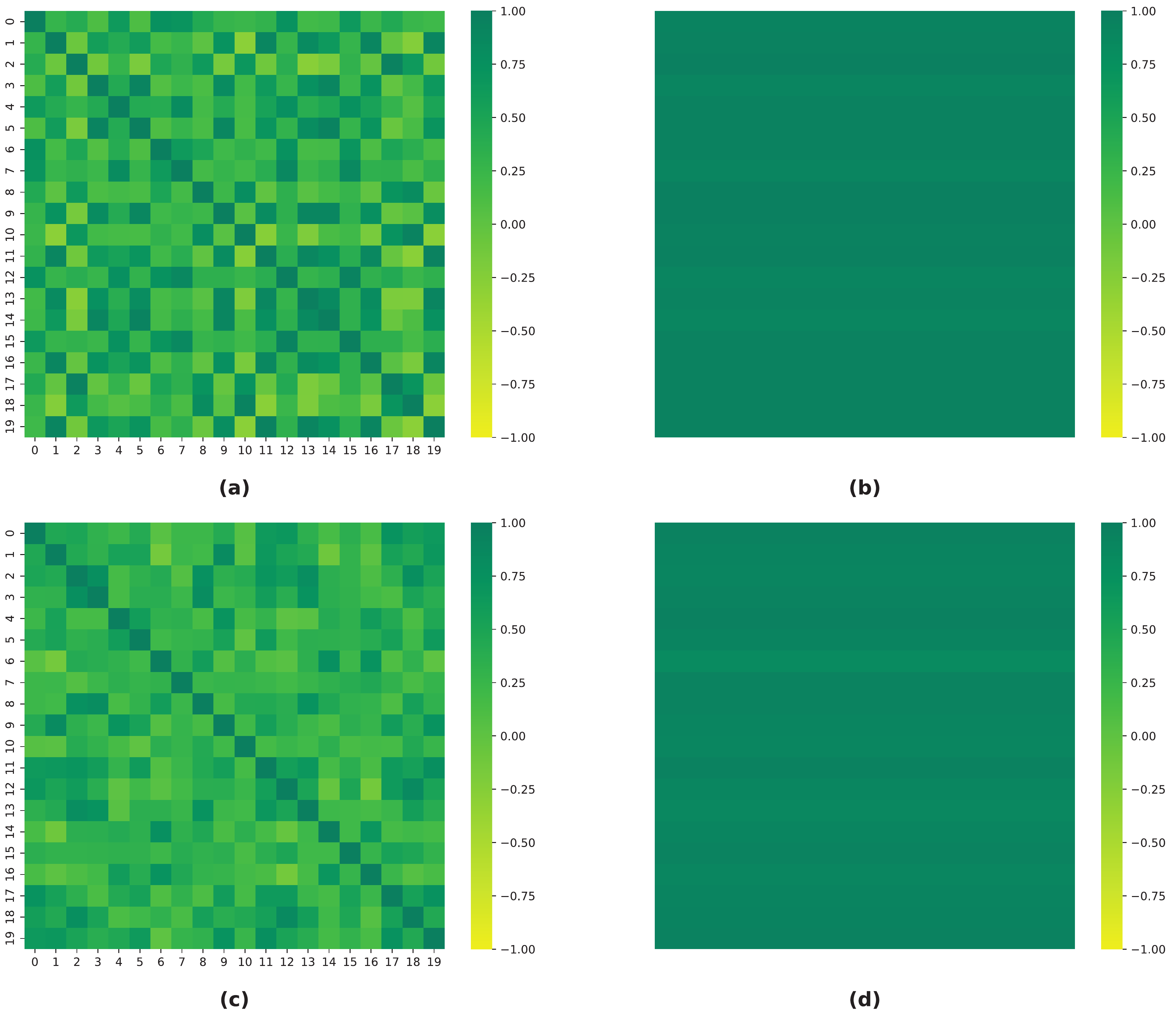}
\caption{Heat map for rule representation similarity. (a) Inter-problem rule representation similarity for `Center' configuration. (b) Intra-problem rule representation similarity for `Center' configuration. (c) Inter-problem rule representation similarity for `3 $\times$ 3 Grid' configuration. (d) Intra-problem rule representation similarity for `3 $\times$ 3 Grid' configuration.}
\label{Heat map}
\end{figure}

Recall that RS-CNN evaluates the cosine similarity between the rule representations. Following this logic, it is assumed that rule representations for the same rule, through out the whole dataset, should be identically the same, at least bear extremely close resemblance to each other. However that is not happening in RS-CNN. Take the configuration of `Center' and `3 $\times$ 3 Grid' in RAVEN for example, 20 problem matrices with identical rules are generated for each configuration. As shown in Fig. \ref{Heat map}, for each configuration, although rules for these 20 problem matrices are identical, perfect match of rule representations can only be observed in the intra-problem matrix situation. That is, RS-CNN does not summarize rules in the way predefined in the dataset. On the other hand, RS-TRAN with pre-training aligns rule representations of problem matrices who share the same rule, so as to obtain better performance in general.

\section{Conclusion}
In this paper, we develop two RPM solvers based on CNN and ViT, showing in experiment level that well-designed inductive bias with data compatibility is the key to successful end-to-end RPM neural solvers. Our work also reveals that, meta-data containing non-pixel level information (e.g. rule information, structure information) is not a `must-have' in abstract reasoning, given that sufficient pixel-level data with full coverage of rules is provided. However, the meta-data in proper form is able to accelerate the training process, and improves the performance of model remarkably.

\bibliographystyle{unsrt}

\end{document}